\documentclass[runningheads]{llncs}
\usepackage{graphicx}
\usepackage{subcaption}
\usepackage{float}
\usepackage{dirtytalk}
\usepackage{graphicx}
\usepackage{amsmath}
\usepackage{tabularx,booktabs,caption}
\newcolumntype{Z}{>{\raggedright}X}
\newcommand*{\Scale}[2][4]{\scalebox{#1}{\ensuremath{#2}}}%
\usepackage{multirow}

\begin{document}
\title{Learning More for Free - A Multi Task Learning Approach for Improved Pathology Classification in Capsule Endoscopy}
%
%
\author{Anuja Vats\inst{1}\and 
Marius Pedersen\inst{1}\and
Ahmed Mohammed\inst{1,2} \and
Øistein Hovde \inst{1,3,4}}
%
\authorrunning{A. Vats et al}
%
\institute{NTNU, Gjøvik, Norway 
\email{\{anuja.vats,marius.pedersen,mohammed.kedir\}@ntnu.no}\\ \and
SINTEF Digital, Smart Sensor Systems, Oslo, Norway  \and
University of Oslo, Norway \and
Innlandet Hospital Trust, Gjøvik, Norway 
\email{oistein.hovde@sykehuset-innlandet.no}}
\maketitle              
\begin{abstract}
The progress in Computer Aided Diagnosis (CADx) of Wireless Capsule Endoscopy (WCE) is thwarted by the lack of data. The inadequacy in richly representative healthy and abnormal conditions results in isolated analyses of pathologies, that can not handle realistic multi-pathology scenarios. In this work, we explore how to learn more for free, from limited data through solving a WCE multicentric, multi-pathology classification problem. Learning more implies to learning more than full supervision would allow with the same data. This is done by combining self supervision with full supervision, under multi task learning. Additionally, we draw inspiration from the Human Visual System (HVS) in designing self supervision tasks and investigate if seemingly ineffectual signals within the data itself can be exploited to gain performance, if so, which signals would be better than others. Further, we present our analysis of the high level features as a stepping stone towards more robust multi-pathology CADx in WCE. Code accompanying this work will be made available on github.
\end{abstract}
\section{Introduction}
WCE is a diagnostic procedure for screening the lining (mucosa) of the Gastrointestinal Tract (GIT) for abnormalities. It has succeeded in being the first line of diagnosis for the middle part of the GIT \cite{yang2020future}. Apart from being widely preferred for examining the small bowel, due to its non-invasive nature, it is also a preferred alternative to upper endoscopy and colonoscopy \cite{mohammed2018stochastic,park2018eso}.

The procedure involves a pill-sized imaging device shaped as a capsule, to be swallowed by the patient. It traverses through the GIT taking pictures of the mucosa and lumen, transmitting them wirelessly to an external data recorder. The data from each capsule is a video stream that is 8 to 16 hours long, depending on the capsule model \cite{mcalindon2016capsule,laiz2019using,atsawarungruangkit2020intro} amounting to frames upwards from 50,000 to be investigated by experts. Moreover, a significant number of these frames contribute from little to no diagnostic information due to either debris and bubbles completely or partially occluding the mucosa or undesirable capsule orientations for long duration during its traversal \cite{segui2016generic}.

Owing to its success in recognition tasks, deep learning has led to major breakthroughs in anomaly detection across medical domains. Soffer et al. \cite{soffer2020deep} present a comprehensive review including deep learning based studies in WCE. Most recent approaches apply a Convolutional Neural Net(CNN) based feature extractor for\textbf{ C}omputer\textbf{A}ided \textbf{De}tection (CADe) of pathologies like polyps, ulcers, bleeding etc.\cite{soffer2020deep,yang2020future,hwang2018application,muhammad2020vision}. 
A multicentric CADe system trained potentially from  millions of images as in \cite{ding2019gastroenterologist} can achieve an even higher detection rate than human-readers for certain anomalies. They perform the first of such a large scale multicentric study by rigorously fine-tuning a pre-trained CNN on more than 100 million images. While their method achieves high sensitivity in filtering out suspected anomalies, further attempt to classify the detected anomaly is not as  successful. Considering the large amount of multicentre data used in this study, the inability to do so, reveals the complexity involved in reliably representing features that can classify multiple abnormalities beyond obvious confounders. Infact, this happens to be one of the key challenges in CADx arising from the lack of morphological descriptions of pathological and healthy conditions at disposal (the essential difference lies in CADx focusing on features for the purpose of characterizing an abnormality as opposed to localizing an abnormality within a frame (CADe)). Stidham et al. \cite{syed2020histopatho} discuss the consequences of such a challenge, with regards to tissue histology. Some of the complicating factors can be understood by considering the response of the GIT under abnormality:
\renewcommand\labelitemi{$\cdot$}
\begin{enumerate}
    \item It may exhibit similarity in visual characteristics for different conditions and different severity of these conditions.
    \item It may exhibit visible differences in pathology appearances that may get overshadowed by other (medically irrelevant but visually pervasive) morphological similarities pertaining to the endoluminal scene (e.g. mucosal surface structures, occlusions, turbidity, illumination etc.)
\end{enumerate}

Our strategy to combat this challenge is to learn more efficiently discriminative features with limited data for free. We achieve this by taking advantage of additional (seemingly ineffectual) teaching signals within data through two ways - self supervision and Multi Task Learning (MTL). The inspiration behind MTL comes from the observation that humans greatly benefit, in their learning, by performing multiple tasks as opposed to learning the same tasks in isolation \cite{Caruana93multitasklearning}. By the same logic, neural networks benefit by sharing information across several related tasks, by means of a common underlying representation. Even when the primary focus is only on one of the tasks, a good choice of additional tasks can lead to better predictive performance by supplementing domain-specific information not necessarily captured by the main task \cite{Caruana93multitasklearning,caruana1998multitask,zhang2014facial,liu-etal-2015-representation,misra2016cross,benton2017multitask}. 

Notwithstanding the challenges of WCE, gastroenterologists are usually able to find pathological areas within the image with relative ease. Most lesions and inflammations are obscure and vary considerably in scale considering the global context within images. In addition, the inherent presence of distortions causes further corruption of diagnostically relevant details. Despite this, gastroenterologists can not only identify and localize abnormality but also classify its type. That is, aside from the most difficult of cases (\cite{zheng2012disappointing}) the doctor's ability to identify and classify pathologies remains relatively unhindered by distortions. This applies generally to the HVS.
Continuing from this argument, if distortions are unavoidable in WCE and doctors can diagnose in spite of them, can they be exploited to provide a useful inductive bias for classification? To this end, we propose a novel MTL framework as a combination of two types of tasks - supervised pathology classification task (SPT) and  self-supervised (SS) distortion level classification task (SSDT). The SSDT task is SS as the labels are freely obtained from distorting the images in a pre-calculated way to enable supervised learning from it. Further, we also present feature analysis for WCE CAD-CAP data \cite{leenhardt2020giana}. To the best of our knowledge, we are the first ones to perform multicentric pathology classification on WCE using MTL and SS. Besides ours, the same classification has been presented in only one other work \cite{valerio2019GIANA}, however, their objectives and approach differ from ours significantly. The main contributions of this paper are
\begin{enumerate}
    \item We are the first to apply MTL for WCE classification. We combine teaching signals from two different levels of supervision - full supervision (SPT) and self supervision (SSDT) to gain in performance. 
     \item We increase the pathology classification accuracy by 7\% from a Single Task (ST) based pathology classification.
    \item We explore the influence of SSDTs and discuss their role in performance.
    \item We discover that the clusters of similarity that automatically emerge in a high level feature space correspond to those manually identified in other works \cite{segui2016generic,laiz2019using}. This understanding of the similarity explains the cause for the bottleneck of CADx typical in WCE.

\end{enumerate}

\section{Method  } 
\begin{figure}[t]
\centering
\noindent\makebox[\textwidth]{\includegraphics[width=\textwidth]{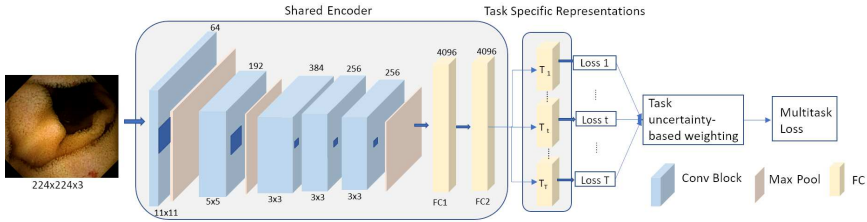}}
\caption{Overview of our proposed MTL architecture. The encoder holds a shared representation that is simultaneously discriminative for multiple tasks, whereas the task specific layer specializes in a particular task. }
\label{Arch}
\end{figure}

Major distortions in WCE are similar but aggravated from those identified in clinical endoscopy \cite{ali2019EAD}, owing to the uncontrolled motion of the capsule inside the GIT. In this work, three distortions have been selected for auxiliary SSDT - brightness (B), contrast (C) and motion blur (MB). One MTL experiment is a combination of SPT with one (or two) SSDT. The SSDT comprises of adding a known level of the selected distortion to be predicted onto a centrally cropped, resized (400x400) and randomly rotated image. Each level of distortion (numerical value between (0,3]) corresponds to an output category for SSDT B and C (for more details refer suppl. Table 2). For SSDT MB, we choose to add linear motion blur, with output categories corresponding to the direction of added motion [0,45,90,180]. The network at any time does two simultaneous- 3 way-pathology and 4 way-distortion level classifications. The training scheme employed is random network initialization with hard parameter sharing \cite{Caruana93multitasklearning}, under which all layers upto the penultimate task-specific layers share the same representation. This is especially useful when the training data is limited. In having to generalize for multiple tasks using one representation, the network is less prone to overfitting~\cite{baxter1997bayesian}. Fig. \ref{Arch} illustrates the architecture, which has been adapted from AlexNet \cite{krizhevsky2017imagenet}. Five Convolutional (Conv) and two Fully Connected layers (FC1, FC2) are shared across the tasks whereas, the final FC layers are task specific. A single input image contains signal for both classification tasks at any time.

\subsubsection{Objective Function -} Every MTL is a joint optimization of two or more losses. In this problem, one of the losses corresponds to SPT while other to SSDT. The total objective conventionally is a weighted sum of individual objectives, in this case a cross-entropy objective for each task. Let $ D^{t} = (x_{i}, y_{i}^{t}), i \in {1, 2, ..., N}$ represent the dataset for a task $t$ obtained from a set of tasks given by $t \in {1, 2, ..., T}$. $ (x_{i}, y_{i}^{t}) $ represent the input output pair for task $t$ such that, in our scheme, each image $ x_{i}$ contains the training signal for all the tasks, and $y_{i}^{t}$ is the corresponding task label vector. If $\mathcal{L}_{cet}$ is the cross-entropy loss for task $t$, then a conventional formulation can be given by

\begin{equation}
\Scale[1.1]
{ \underset{\theta}{\text{argmin}} \sum_{t=1}^{T} \lambda_{t}\mathcal{L}_{cet}(\theta, f(X^{t}), Y^{t}) + \alpha \Omega(\theta) }
\end{equation}

where $f(X^{t})$ is the prediction for task $t$, $\theta$ are the network parameters and $\lambda$ is the task-specific weight. The term $\alpha \Omega(\theta)$ corresponds to regularization concerning network weights with $\alpha \in [0, inf) $. This formulation assumes that optimal weighting of tasks $\lambda_{t}$ is either known or can be found. In practice it requires manual tuning which is unintuitive and expensive. We adopt the approach proposed by Kendall et al.\cite{kendall2018uncertainity}, where the contribution towards the total loss by a task is learned on the basis of homoscedastic uncertainty. This extends previous formulation for multiple classification tasks as

\begin{equation}
\Scale[1.1]
{\underset{(\theta, \sigma)}{\text{argmin}}\sum_{t=1}^{T}\Bigg(  \frac{1}{\sigma_{t}^2}\mathcal{L}_{cet}(\theta, f(X^{t}), Y^{t}) + log \sigma_{t}\Bigg) + \alpha \Omega(\theta)}
\end{equation}
where $\frac{1}{\sigma_{t}^{2}}$, the relative contribution of a task to the loss, is an indirect estimate of the uncertainty of task $t$ and learning  $\sigma_{t}$ can be interpreted as learning the temperature of a Boltzmann distribution \cite{kendall2018uncertainity}. As a task with high uncertainty begins contributing more towards the total loss compared to a task with lower uncertainty, the relative contribution and eventually task uncertainties themselves are optimized as training progresses, that is to say, the optimization in Eq.(2) with respect to both network parameters $\theta$ along with $ \sigma_{t}$, allows the relative contribution of each task towards total loss to change during the course of training till a reasonably low ratio is reached for all tasks involved. The additional term $log \sigma_{t}$ is an additional regularizer, penalizing the tendency of $\sigma_{t}$ towards larger values.

\section{Experiments and Results}
The data used for our multicentric pathology classification is from the Computer Assisted Diagnosis for Capsule Endoscopy Database CAD-CAP (GIANA Endoscopic Vision Challenge 2018) comprising of 20,000 normal and 5000 images of varying abnormalities \cite{leenhardt2020giana}. Out of these, a balanced classification dataset with 1812 images belonging to three classes \say{normal}, \say{inflammatory lesion} and \say{vascular lesion} (600, 607, 605 respectively) has been used. 80:10:10 split has been used for train, validation and test sets. In the challenge, the dataset has been used for CADe, while in this paper we use it for CADx. The five MTL configurations studied in this paper are - STL : fully supervised pathology classification (only SPT), MTL (P+X) : refers to (SPT + SSDT of X), where,
SSDT of X is the SS auxiliary distortion task of distortion X. X could be motion blur (MB), Brightness (B) or contrast (C). MTL (P+X+Y) is a three task configuration with SPT and a chosen combination of two SSDT of X. 
We train using Adam optimizer for 300 epochs with a batch size of 64, an initial learning rate of 0.01 and decay rate of 0.1 every 50 epochs on Nvidia Twin Titan RTX. 
\subsubsection{\textbf{Results and Analysis -}} Table~\ref{results} shows the results of our experiments. Overall MTL (P+B) achieves the highest accuracy, but on account of maximizing the correct classification of normals, with no improvement (in fact worse discrimination as compared to STL) for the pathology classes.  MTL (P+C) is the best with respect to pathology sensitivity, i.e. while the classification of normals is only slightly better than STL, change in image contrast enables pathologies to be better discriminated. MTL (P+MB) is marginally better than STL.

\begin{table}[h!]
\caption{Performance of STL and different MTL configurations. $P_{i}$ and $P_{v}$ refer to the pathology class of inflammatory and vascular lesions respectively. N refers to the class of normals.}\label{results}
\setlength\tabcolsep{8pt} 
\begin{tabularx}{\columnwidth}{@{} Z *{9}{c} @{}}
\toprule 
Config   & \multicolumn{2}{c@{}}{Accuracy (\%)} & \multicolumn{3}{c@{}}{Sensitivity (SPT)} & \multicolumn{3}{c@{}}{Specificity (SPT)} \\
\cmidrule(lr){2-3} \cmidrule(lr){4-6} \cmidrule(lr){7-9}
& Pathology & Distortion  &  $P_{i}$ & N & $P_{v}$ &  $P_{i}$ & N & $P_{v}$ \\
\midrule 
STL    & 51.9 & --    &    0.44 &  0.72  & 0.44   &    0.41 &  0.68  & 0.47 \\ 
\midrule 
P+MB     & 52.1 & 40    &    0.45 &  0.70  &  0.38 &  0.49 &  0.57  & 0.44 \\
P+C     & 57.1 & 80.95    &     \textbf{0.56} &  0.75  &  \textbf{0.48} &    0.55 &  0.68  & 0.53 \\
P+B     & \textbf{58.7} & 85.7    &    0.42 &  \textbf{0.88} & 0.35 & 0.46 &  0.74  & 0.39 \\
\midrule 
P+B+C     & 57  & 73.5, 88.8    & 0.55    &0.79  & 0.47 & 0.59    &0.55  & 0.39\\ 
\bottomrule 
\end{tabularx}
\end{table}

We observed that combining more than one SSDT with SPT (MTL P+B+C) does not further improve the overall performance, (Table \ref{results}, supp. Fig 5). This result is not unexpected even when combining fully supervised tasks \cite{kokkinos2017ubernet}. However, compared to STL the gain by combining two tasks comes both in terms of overall accuracy (from correctly classifying normals as in the brightness task) and discriminating between pathologies (higher sensitivity from the contrast task). That is, this task combination can be seen as a way to gain a specific advantage from each of the two auxiliary SSDTs.

\begin{figure}[htp]
\centering
\noindent\makebox[\textwidth]{\includegraphics[width=\textwidth]{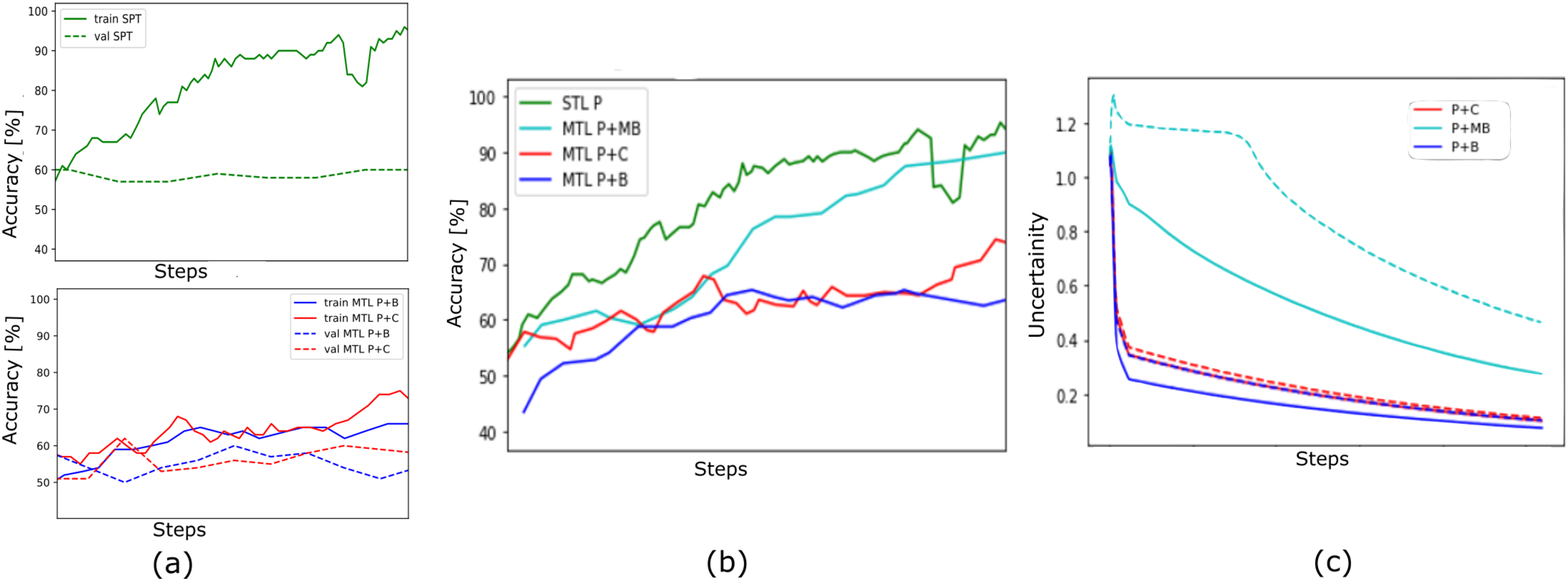}}
\caption{Impact of SSDTs on the training performance of SPT. (a-top) Overfitting in SPT when trained as a supervised task. (a-bottom, b) Training of SPT under MTL, reduced overfitting with longer training time (c) Square-root of task-uncertainities $\frac{1}{\sigma_{p}^2}$, $\frac{1}{\sigma_{d}^2}$ given by $\frac{1}{\sigma_{p}} (solid)$, $\frac{1}{\sigma_{d}} (dashed)$ is shown in plot.} 
\label{2_MTL}
\end{figure}

Further, we analyse how different MTL configurations impact the learning of the pathology task (SPT). Fig. \ref{2_MTL} (a, b) show the training accuracy of SPT with and without MTL, (c) the plots for $\frac{1}{\sigma_{t}}$. 

It can be seen that, in isolation SPT has a high tendency towards over-fitting with quickly escalating training accuracy but plateauing validation accuracy (Fig. \ref{2_MTL} (a)). The addition of an auxiliary SSDT is seen to reduce this effect (Fig. \ref{2_MTL} (a,b)). This is a well known advantage of MTL \cite{baxter1997bayesian} as an overall regularizer. This effect is directly linked to performance seen before, with brightness, contrast and motion blur in order of decreasing overall accuracy. This is also correlated with the trend in uncertainties. The nature of helpfulness of a task towards pathology can be inferred from how the uncertainties behave during training. From  Fig. \ref{2_MTL} (c), for a helpful SSDT, both uncertainties for SPT ($\frac{1}{\sigma_{p}}$) and SSDT ($\frac{1}{\sigma_{d}}$) reduce in tandem, over the course of iterations as more and more confidence is gained from the task. In a relatively unhelpful task (MTL P+MB) however, higher uncertainties indicate this lack of confidence in the task, towards outputs. More specifically, choosing an auxiliary task more complicated (MB) than the main task (SPT) (as seen from the low distortion accuracy for MB in Table~\ref{results}) can lead from little to no benefit from MTL. This complexity arising in the blur task may be due to the varying amount of inherent linear, rotational and motion blur in the images, that may make prediction of the exact level, especially difficult.

\begin{figure}[t]
\centering
\noindent\makebox[\textwidth]{\includegraphics[width=\textwidth]{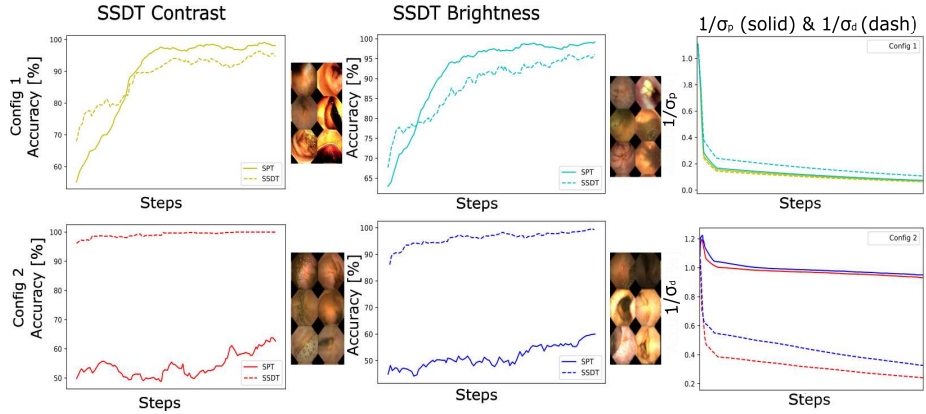}}
\caption{Comparison of two different SSDT designs. Config 1. improves SPT beyond STL performance and vice versa for Config 2. Pre-processed images, adjacent to the graphs illustrate the distortion levels for the two configs.} 
\label{compare}
\end{figure}

We further investigate the design consideration for an already beneficial task to remain beneficial. Fig. \ref{compare} shows the training accuracy for two different design choices for SSDT C and B.
Config 1 which results in better performance for both tasks is a task of higher complexity (smaller difference between levels of distortion, for this experiment this difference was found to be 0.1 or less for both SSDT B and C) compared to Config 2 (larger difference between levels, 0.3 or more)  that results in poorer performance (no gain from STL). This may seem to be in contradiction to the observation before, however it is not. For an MTL to be beneficial the auxiliary task must neither be too easy nor too hard, as in either case, a plateauing effect is observed. MTL P+MB is a difficult task even for simpler designs (larger kernel size being simpler design as the added blur becomes increasingly easier to detect with larger kernels, and vice versa), however, for SSDT C and SSDT B which are tasks of lower complexity with regards to pathology, the plateauing occurs when they are made simple by design. This is done by increasing discrimination between the levels that correspond to output categories (refer Fig.\ref{compare}, suppl. Table \ref{table:x}), In other words, since the auxiliary task is too easy, the distortion task has very high accuracy from the very beginning of the training process. Coupled with a difficult task (SPT), an easy SSDT loses its primary advantage of helping the pathology task out of early local minimas. These findings regarding the nature of beneficial tasks coincide with the observation from~\cite{bingel2017identifying}. Fig.\ref{compare}(c) shows the task uncertainties during training.

\subsubsection{A note on similarity} - In this section, our focus is to answer the question about the correspondence between  \textit{representational similarity} and \textit{medically-perceived similarity}. Medical similarity can be simply described as - two images having the same pathology are similar and different from all images with different pathology. Representational similarity on the other hand is the similarity as understood by the network on the basis of extracted features. A disagreement between these two reveals the cause for confusion in the network. Although MTL is less prone to this than STL, it still is a major bottleneck in performance. We projected the features from FC2 (Fig. \ref{Arch}) on 3D after dimension reduction using Uniform Manifold Approximation and Projection for Dimension Reduction (UMAP) \cite{mcinnes2018umap}. UMAP helps us interpret not only, what features dictate which images are most similar with respect to each other (local), but also what two features are more similar compared to others (global).  

\begin{figure}[t]
\centering
\noindent\makebox[\textwidth]{\includegraphics[width=\textwidth]{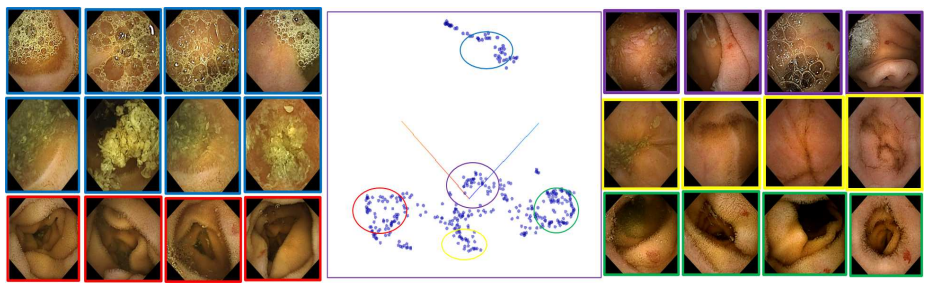}}
\caption{UMAP of features from FC2 in MTL P+C. Each set of neighbors reveal a dominant similarity. Bubbles and turbid (blue),mucosal folds (red),clear normal (yellow),lumen (green),clear pathology (violet)}
\label{features}
\end{figure}

As seen in Fig.\ref{features}, MTL reveals the inherent variations within the dataset. The extracted features are such that while images with the same pathology lie close (even when in different clusters), pathology does not dictate similarity everywhere. The images within a cluster have representational similarity, that may be medically unimportant. In this work, we saw five such dominant factors emerge out of the feature space repeatedly - bubbles and turbidity, folds in mucosa, lumenal view, clear normal, clear pathology (these were verified by a medical expert to be similarities typical to endoluminal structure or scene, except for few clusters of clear pathology where pathology was the dominant feature). Moreover, at global scale, the feature \say{bubbles and turbidity} appears more dissimilar from all other features in all experiments. Images in between clusters show mingled characteristics. Similar dominant features have been used by other authors in works before \cite{segui2016generic,laiz2019using}. We stress at this point that, unlike other works, these features emerge automatically for us as a result of the learning strategy applied and training under STL does not provide this advantage (refer suppl Fig 6. for more details).
 
\section{Conclusion}
In this paper, we demonstrate an MTL approach to gain performance over supervised WCE classification with no added data and a simple CNN architecture, purely by exploiting signals inherent in data under self supervision, additionally to supervision. Further, forcing the network into sharing features between multiple tasks reveals a hierarchy of features in WCE whereby factors that dictate similarity are not inspired only by the presence or absence of pathology, rather are typical of the endoluminal scene. Hence, images with different pathology can be easily perceived to be similar in the presence of such a factor. Our method can be used as a filter for these irrelevant salient patterns, allowing a better subsequent characterization of pathology in CADx. We also experimentally correlate task uncertainties with the nature (beneficial and non beneficial) of tasks \cite{bingel2017identifying} which can be used as a way to choose good auxiliary tasks for MTL. This framework is easily scalable to more complex architectures and auxiliary tasks. 

%
%
%
 \bibliographystyle{splncs04}
 \bibliography{refrences}
%





\section{Supplementary }
\begin{figure}[!ht]
\centering
  \includegraphics[width=\textwidth]{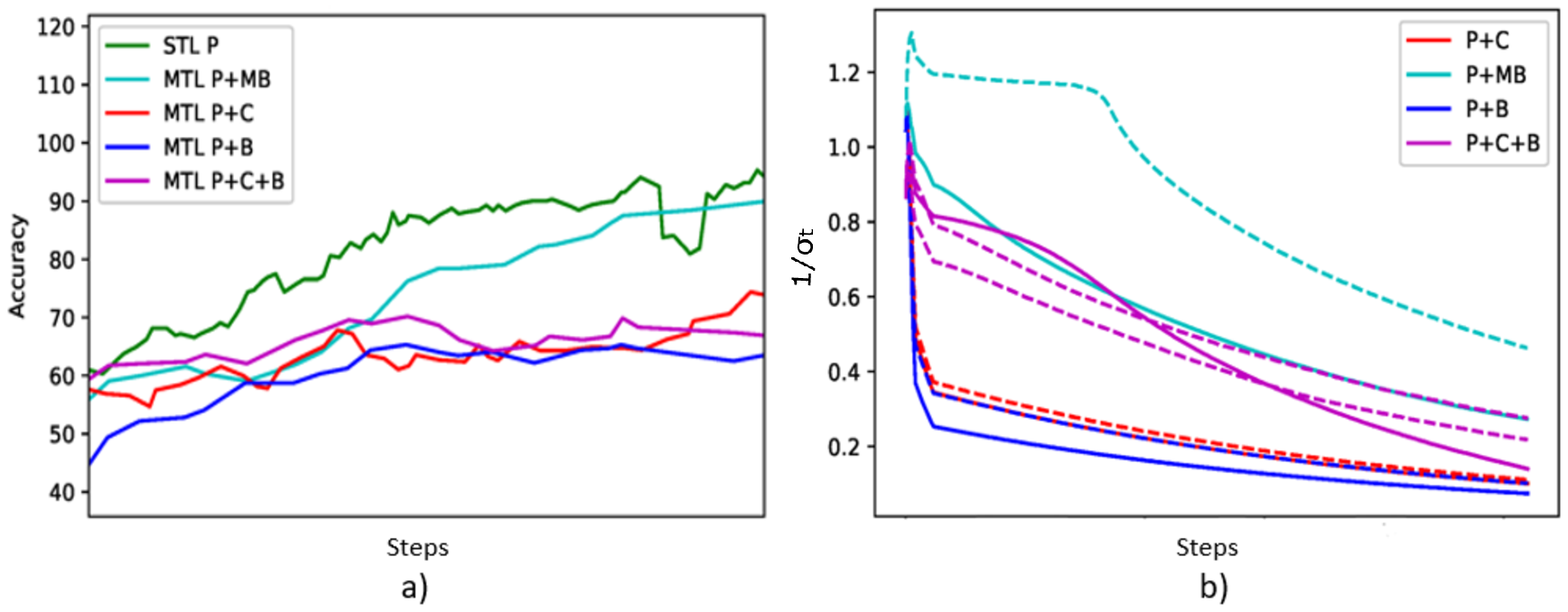}
  \caption{Training Accuracy and Task uncertainty ($\sigma_{t}$) for all MTL configurations. P+X+Y remains close in performance but not above P+X. In [12] combining three tasks improves performance by almost 1\%, however high relatedness can be established between their three tasks (semantic, instance and depth segmentation) and resulting features are still beneficial to all tasks. In our method, since the target is to exploit freely available signals, it's possible that combining two different distortion tasks violates this premise e.g, scale of features (more global (SSDT) than local (pathology)). [13] who also see drop in combining tasks suggest hyper-parameter tuning remedies that can be explored further.}
  \label{3mtl}
\end{figure}
\begin{figure}[!ht]
\centering
  \includegraphics[width=\textwidth]{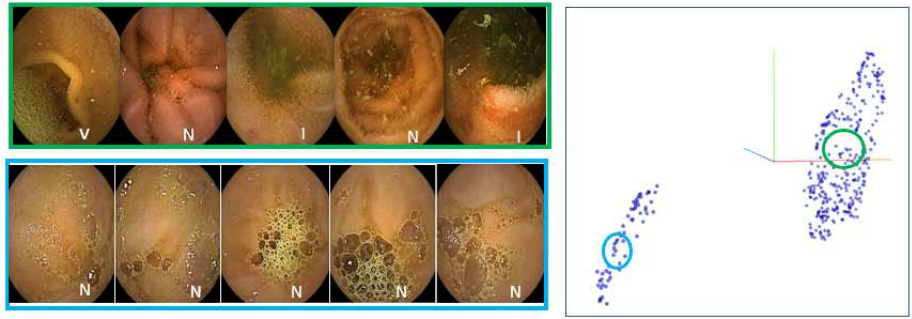}
  \caption{UMAP of features from FC2 in STL. Unlike Fig. 4 (main paper), STL shows largely only two clusters. One corresponds to normals with occlusions (bubbles and debris), while the other consists of all the remaining images (with clean normals), this happens because, bubbles and debris are confounders for 'Normal' class. We observed this as a result of slight bias in the dataset, where more bubbles and debris is present in the normal images as compared to in other classes, as a result of which the network cheats by identifying bubbles instead of characterizing normal mucosa. This effect (though present) is reduced in MTL  as seen by an increase in sensitivity for the normal class through SSDT B and C (Table 1). }
  \label{UMAP nONE}
\end{figure}

\begin{figure}[!ht]
\centering
  \includegraphics[width=\textwidth]{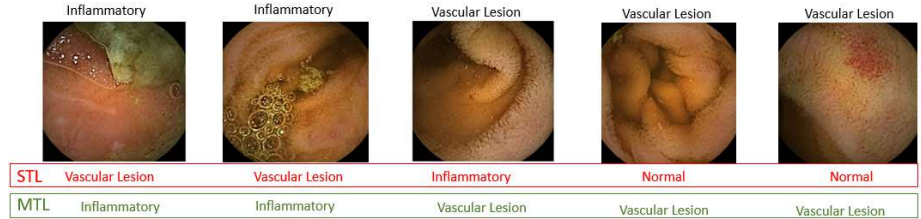}
  \caption{Better discrimination in MTL - Images misclassified in STL(predicted category in red) but correctly classified in MTL P+C (green). In such cases, despite a dominant representational similarity, better discriminative features in MTL allow even obscure pathologies (image 2,3 and 4) to be identified.}
  \label{}
\end{figure}

\begin{figure}[H]
\centering
  \includegraphics[width=\textwidth]{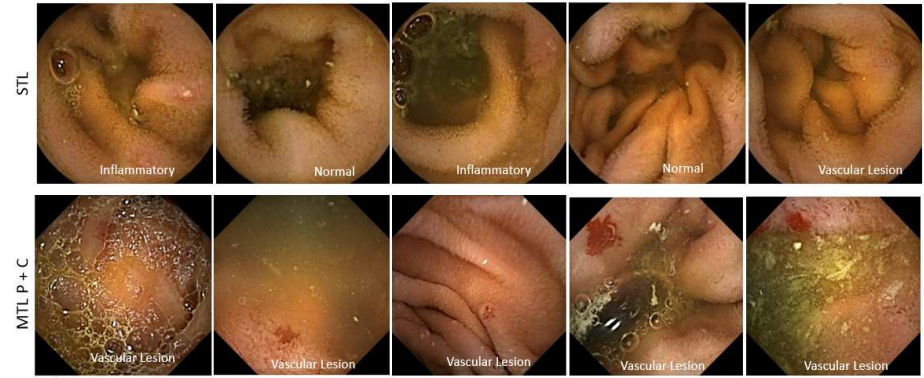}
  \caption{The reduced effect of confounders in MTL. Figure shows the nearest neighbors in space, in STL (row 1) the typical small bowel mucosa with folds and villi lead to different pathologies being perceived similar, however, in MTL this undesired effect is reduced (row 2). Images with varying features (bubbles, folds, debris) lie in the vicinity of each other due to the presence of vascular lesion.}
  \label{}
\end{figure}
\begin{table}[t]
\centering
\caption{Changing between configuration involves making the task easy or hard. This is achieved by changing the differences between the numerical values of the applied distortion for SSDT B and C or changing the kernel size for SSDT MB. The table below shows example categories for each configuration.}
\begin{tabular}{|c|c|c|c|c|c|} 
\hline
Distortion & Design choice & Config & levels & Criteria & Library \\ 
\hline
\multirow{2}{*}{B,C}&\multirow{2}{*}{ difference between levels} & Config 1&{[0.8,0.9,1,1.1]}& 0.1 or less & PIL\\
 &   & Config 2 &{[0.7,1,1.3,1.6]} & 0.3 or more &PIL\\
 
Motion Blur & kernel size & always Config 2 & 3, 5, 10, 15& &  opencv \\
\hline 

\hline
\end{tabular}
\label{table:x}
\end{table}

\end{document}